\documentclass[10pt,twocolumn,letterpaper]{article}

\usepackage{authblk}
\usepackage{iccv}
\usepackage{times}
\usepackage{epsfig}
\usepackage{graphicx}
\usepackage{amsmath}
\usepackage{amssymb}
\usepackage{subfigure}

\usepackage{color, colortbl}

\usepackage{booktabs}
\usepackage{multirow}

\usepackage[accsupp]{axessibility}
\usepackage{eso-pic}

\usepackage[breaklinks=true,bookmarks=false]{hyperref}

\usepackage{caption}
\usepackage{hyperref}

\iccvfinalcopy 

\def\iccvPaperID{****} 

\ificcvfinal\fi

\begin{document}

\title{DyGait: Exploiting Dynamic Representations for \\ High-performance Gait Recognition}


\author[1]{Ming Wang}
\author[2]{Xianda Guo \textsuperscript{$\dagger$}}
\author[3]{Beibei Lin}
\author[2]{Tian Yang}
\author[2]{Zheng Zhu}
\author[4]{Lincheng Li}
\author[1]{Shunli Zhang \thanks{Shunli Zhang is the corresponding author. \par
$\dagger$ Joint first authors. }}
\author[5]{Xin Yu}
\affil[1]{Beijing Jiaotong University} 
\affil[2]{PhiGent Robotics}
\affil[3]{National University of Singapore}
\affil[4]{NetEase Fuxi AI Lab}
\affil[5]{University of Technology Sydney}


\maketitle
 \ificcvfinal\thispagestyle{empty}\fi

\begin{abstract}
Gait recognition is a biometric technology that recognizes the identity of humans through their walking patterns. Compared with other biometric technologies, gait recognition is more difficult to disguise and can be applied to the condition of long-distance without the cooperation of subjects. Thus, it has unique potential and wide application for crime prevention and social security. At present, most gait recognition methods directly extract features from the video frames to establish representations. However, these architectures learn representations from different features equally but do not pay enough attention to dynamic features, which refers to a representation of dynamic parts of silhouettes over time (\eg legs). Since dynamic parts of the human body are more informative than other parts (\eg bags) during walking, in this paper, we propose a novel and high-performance framework named DyGait. This is the first framework on gait recognition that is designed to focus on the extraction of dynamic features. Specifically, to take full advantage of the dynamic information, we propose a Dynamic Augmentation Module (DAM), which can automatically establish spatial-temporal feature representations of the dynamic parts of the human body. The experimental results show that our DyGait network outperforms other state-of-the-art gait recognition methods. It achieves an average Rank-1 accuracy of 71.4\% on the GREW dataset, 66.3\% on the Gait3D dataset, 98.4\% on the CASIA-B dataset and 98.3\% on the OU-MVLP dataset. 
\end{abstract}

\section{Introduction}
\label{sec:intro}
Gait recognition is a biometric technology that can identify humans based on their postures at a long distance without the cooperation of subjects. Meanwhile, gait is hard to disguise, and these advantages give gait recognition unique potential for many applications such as crime prevention, person identification, and social security. Even though gait recognition has achieved impressive advances in past years \cite{lin2021gait,lin2020gait,hou2021set,hou2020gait,fan2020gaitpart,chao2019gaitset,huang2022star}, this challenging technique has not yet been widely used in real-world applications due to some complex factors, such as clothing and people carrying objects \cite{connor2018biometric,liao2017pose,yu2006framework,bao2022using,tian2021blind,sarkar2005humanid,yu2021hid,shen2022comprehensive}.

\begin{figure}[t]
\centering
\includegraphics[width=0.98\linewidth]{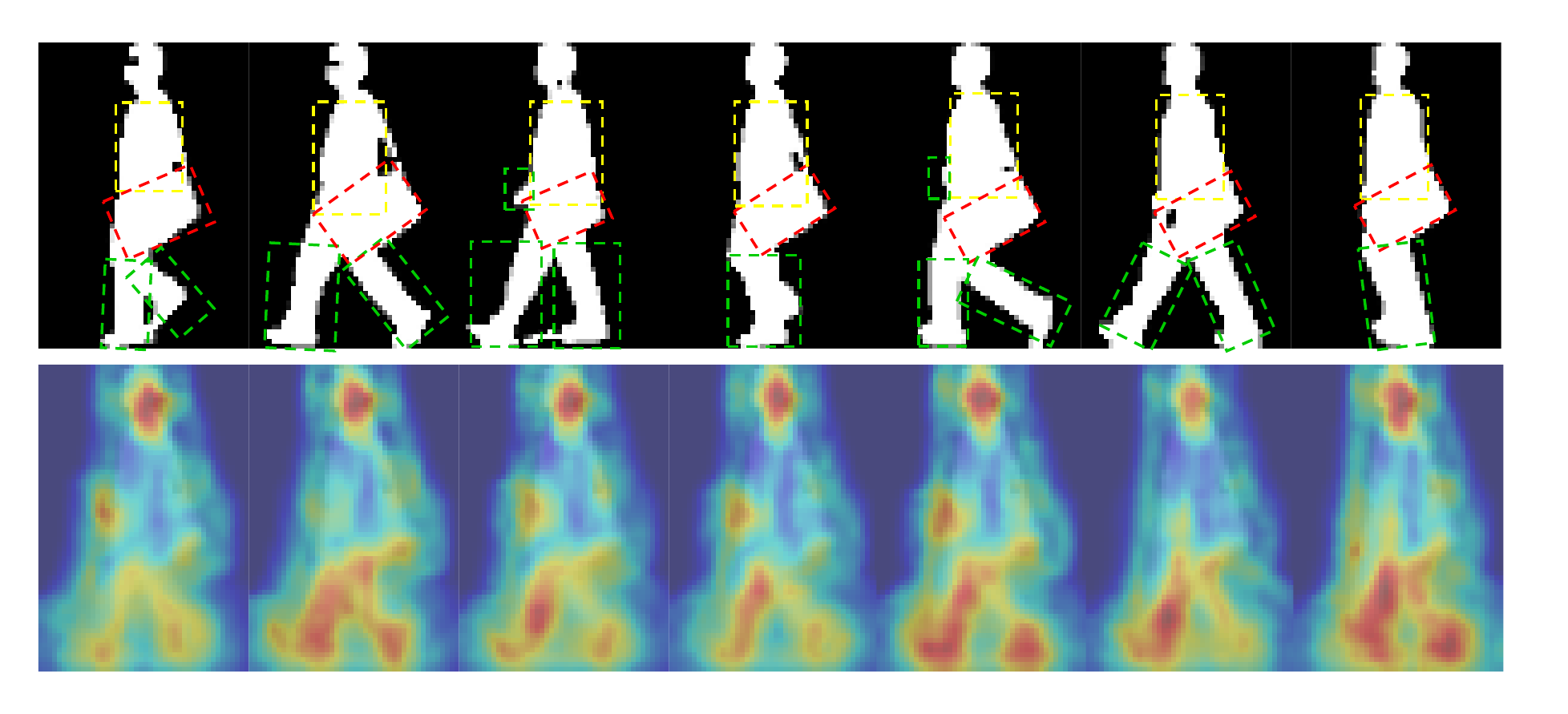}
\caption{The first row represents the original sequence of gait. The bounding boxes for human silhouette parts: main body (yellow boxes), bag (red boxes), left and right legs (green boxes). The second row shows heatmaps of our DyGait.}
\vspace*{-1em}
\label{Fig_Mask}
\end{figure}

As an identification task in computer vision, the essential goal of gait recognition is to learn unique and invariant representations from temporal changing characteristics about the motion of bodies. Currently, there are two main types of feature representation methods to portray human gait. One is the spatial-based method which usually extracts spatial gait features of the whole gait sequences  \cite{chao2019gaitset,GEI}. Despite having a low computational cost in this way, it may lose temporal information. The other one is the temporal-based method which models temporal features for representations \cite{fan2020gaitpart,lin2021gait}. These CNN-based works can automatically extract spatial and temporal gait features, but they do not focus on the dynamic differences between frames, which may be the most useful gait features for gait recognition.

As the first row in Figure \ref{Fig_Mask} shows, the human's torso and the bag occupy a large region in each frame and remain almost static and unchanged during the walking process. In contrast, the legs take up small spaces and are continuously changing in motion. It can be observed that main differences of the gait among different frames lie in the dynamic features such as the moving legs. This suggests that some of the static regions, \eg bag or coat, are not critical to distinguish one person from the others. From this perspective, we stress that the dynamic parts of the human body are more informative than others. 
Therefore, the applied approach should pay more attention to the dynamic feature.

Motivated by the above observation, we propose a novel gait recognition method called DyGait, which can automatically extract dynamic information of the gait. As shown in Figure \ref{Fig_Mask}, DyGait pays more attention to the motion parts, such as legs and arms.

First, a novel Dynamic Augmentation Module (DAM) is developed to extract more comprehensive representations.
DAM is built based on Dynamic Feature Extractor (DFE), which can ensemble the global temporal information of feature maps to generate a gait template. Then, the dynamic feature maps can be obtained by computing the difference between the feature maps of each frame and the gait template. 

In addition, both Temporal Aggregation (TA) and Horizontal Mapping (HM) operations are applied to generate feature representations \cite{lin2021gait}. The proposed DyGait achieves strong performance and outperforms other state-of-the-art models by a large margin on GREW, Gait3D, CASIA-B and OU-MVLP. The main contributions are as follows:

1) We propose a novel framework for gait recognition, called DyGait. To the best of our knowledge, this is the first network that is designed to explicitly focus on the extraction of dynamic features of gait.

2) DyGait is built based on the Dynamic Augmentation Module (DAM), which allows a network to focus on the key information and learn more discriminative representations for gait recognition. Meanwhile, this module can effectively filter invalid noise by paying attention to dynamic information. 

3) We achieve the state-of-the-art performance on the most popular datasets inclduing GREW, Gait3D, CASIA-B and OU-MVLP. It obtains an average Rank-1 accuracy of 71.4\% on GREW, 66.3\% on Gait3D, 98.4\% on CASIA-B and 98.3\% on OU-MVLP, respectively. The experiments demonstrate that our method significantly outperforms the previous methods by a large margin.

\section{Related Work}

\noindent \textbf{Gait Recognition.}
Most prior works \cite{GEI,Wang2012human,wu2016comprehensive} are based on the extraction of handcrafted features from gait sequences using traditional machine learning approaches. Gait energy image (GEI) \cite{GEI} used in such investigations is the most popular approach to describe gait. Although the noise can be effectively suppressed by averaging over the gait cycle within a long temporal range in a GEI, this template loses most details such as temporal information. Inspired by the successful application of Convolutional Neural Networks (CNNs) in face recognition \cite{Deepface,FaceNet,CosFace,ArcFace,Curricularface,CASIA-WebFace,MS1M,MegaFace} and person Re-IDentification (Re-ID) \cite{AlignedReID,BoT,xiao2017margin,hermans2017defense,HPM,zheng2016person,MARS,Market-1501,DukeMTMC,CUHK03,MSMT17}, recent researchers propose many gait recognition frameworks based on CNN. Current works in gait recognition are divided into two types of feature representations: spatial feature representation and temporal modeling.

\noindent \textbf{Spatial Feature Representation.}
The first type regards the gait sequence as a template, which relies on binary human silhouette images. The goal of template generation is to encode a gait cycle into a single image, \ie Gait Energy Image (GEI) \cite{GEI} or a Chrono-Gait Image (CGI) \cite{Wang2012human}. In the template matching procedure, the gait representation is firstly extracted from a template image using machine learning approaches \cite{bashir2010gait,xing2016complete} or deep learning \cite{wu2016comprehensive,shiraga2016geinet,zhang2016siamese,chao2019gaitset,zhang2019cross,zhang2019comprehensive,lin2020gait,chao2021gaitset,chai2021semantically,chai2021silhouette,shen2022lidar}. Then, similarities between pairs of representations are measured using Euclidean distance or other metric learning approaches. For example, Shiraga \etal. \cite{shiraga2016geinet} propose the GEINet framework to extract gait features from Gait Energy Image (GEI), which is generated by using the mean function. Zhang \etal. \cite{zhang2016siamese} also take the GEI as input to extract gait features. However, the generation process of GEI causes severe information loss. Hence, Chao \etal. \cite{chao2019gaitset} propose a GaitSet framework, in which the first step is to extract the static gait features and then use a max function to generate gait templates. Zhang \etal. \cite{zhang2019cross} propose an attention module to learn weights of different frames, and then adopt a weighted average operation to create a gait template. Although these methods can achieve excellent performance and be easy to compute, they do not consider temporal information at the feature extraction stage.

\begin{figure*}[ht]
\centering
\includegraphics[width=1\textwidth]{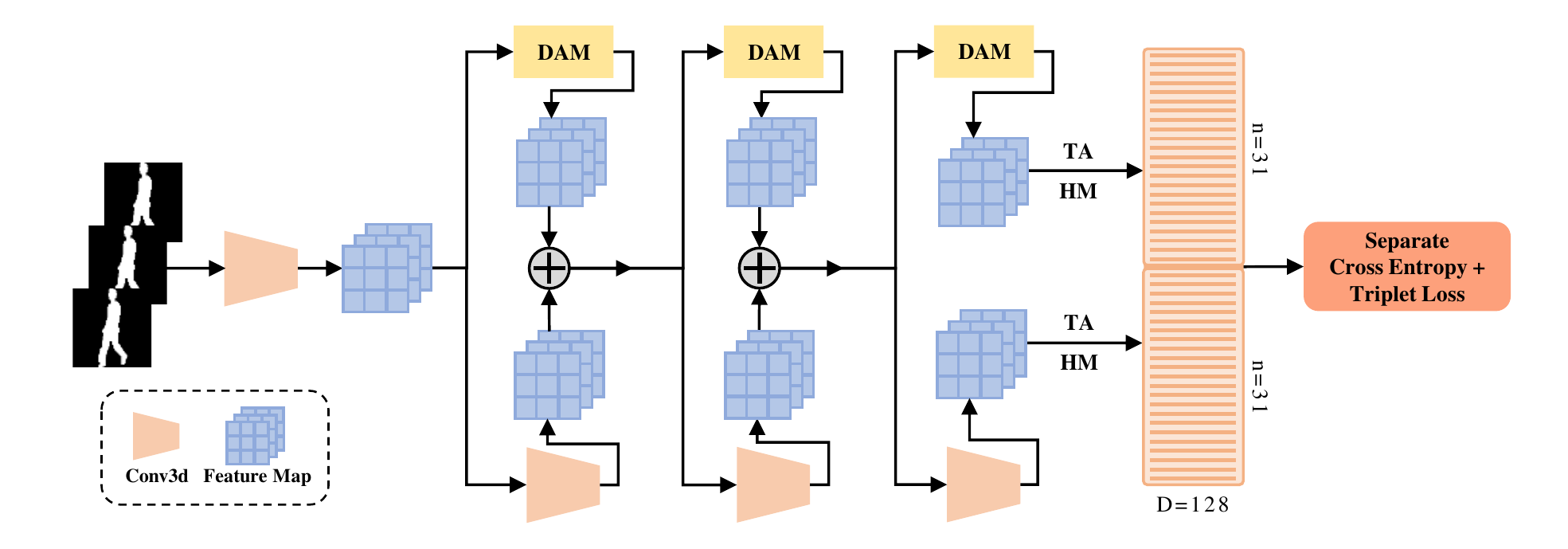}
\caption{Overview of the whole gait recognition framework. HM means Horizontal Mapping. TA denotes Temporal Aggregation. DAM means Dynamic Augmentation Module.
}
\label{overview}
\end{figure*}

\noindent \textbf{Temporal Modeling.}
In the second category, 3D-CNNs \cite{liao2017pose,PoseGait,lin2020gait,lin2021multi,lin2021gaitmask,shen2022gait,wang2022gaitstrip, lin2022uncertainty,dou2022gaitmpl,yu2022cntn,yu2022generalized} or LSTMs \cite{wolf2016multi,sokolova2019pose,thapar2019gait} are used for modeling the temporal information. These approaches can comprehend more spatial information and gather more temporal information but require higher computational costs. Wolf \etal. \cite{wolf2016multi} partition a gait sequence into multiple non-overlapping gait clips and use 3D CNNs to extract each clip's gait features. Thapar \etal. \cite{thapar2019gait} also adopt a similar strategy to extract gait features and further introduce an LSTM module to aggregate multiple clips' features. However, it is inflexible because it only extracts and aggregates a fixed-length clip's information. Recently, Lin \etal. \cite{lin2020gait} propose a novel framework to combine advantages of both template-based and sequence-based methods. They firstly use 3D CNN to extract spatial-temporal gait features and then generate a gait representation by using the statistical function. However, despite the success of spatial feature representation and temporal modeling, their extraction become more complicated for dynamic and changing information. In other words, they do not focus on the most valid information from the gait.

Thus, we turn the attention to dynamic parts of gait and propose the Dynamic Augmentation Module (DAM) which can be used to augment the representation ability. GaitNet proposed by Zhang \etal. \cite{zhang2020learning} is the most related work. Unusually, GaitNet learns a representation of gait directly from RGB frames in videos. Compared with \cite{zhang2020learning}, our approach can automatically disentangle dynamic features from the binary silhouettes, which is beneficial to privacy protection and gives its strong robustness to different clothes/carrying conditions. In addition, some recent work extract gait features through optical flow \cite{sokolova2019pose}, 2D pose \cite{GaitGraph} and 3D pose \cite{PoseGait}. These approaches are robust to clothing variations but depend on the optical flow and pose estimation accuracy.



\section{Proposed Method} \label{method}

In this section, we first overview the framework of the proposed method. Then, we introduce the Dynamic Augmentation Module (DAM), Temporal Aggregation (TA), Horizontal Mapping (HM) and loss function we used. Finally, we explain the training and test details.

\subsection{Overview}
The framework of the proposed method is illustrated in Figure \ref{overview}, which includes Dynamic Augmentation Module (DAM), Temporal Aggregation (TA) and Horizontal Mapping (HM). We firstly use a convolution layer to extract shallow features and then aggregate local temporal information by using the Local Temporal Aggregation (LTA) \cite{lin2021gait}.
Assume that $X_{in}\in \mathbb{R}^{C_{in} \times T_{in} \times H_{in} \times W_{in}}$ is the input gait sequences, where $C_{in}$ is the number of input channels, $T_{in}$ is the length of the gait sequence and ($H_{in}$,$W_{in}$) is the image size of each frame. These operations can be represented as 
\begin{equation}
Y_{L} = \sigma ( C^{3 \times 1 \times 1}( \sigma ( C^{3 \times 3 \times 3}(X_{in}) ) ) ),
\end{equation}
where $Y_{L}\in \mathbb{R}^{C_{L} \times T_{L} \times H_{in} \times W_{in}}$ is the output of the Local Temporal Aggregation (LTA), $C_{L}$ is the number of output channels, and $T_{L}$ is the length of the feature map $Y_{L}$. $C^{3 \times 3 \times 3}$ denotes the 3D convolution with kernel size $3 \times 3 \times 3$. $C^{3 \times 1 \times 1}$ means the 3D convolution with kernel size $3 \times 1 \times 1$ and stride $3$. $\sigma$ means activation function. 

Then, we propose the feature extraction module based on DAM to extract augmented dynamic features. After that, we introduce TA and HM operations to generate feature representations.
Finally, the triplet loss and cross-entropy loss are taken as loss functions to train the proposed network \cite{hou2020gait,lin2021gait}.

\begin{figure}[ht]
\centering
\includegraphics[width=0.85\linewidth]{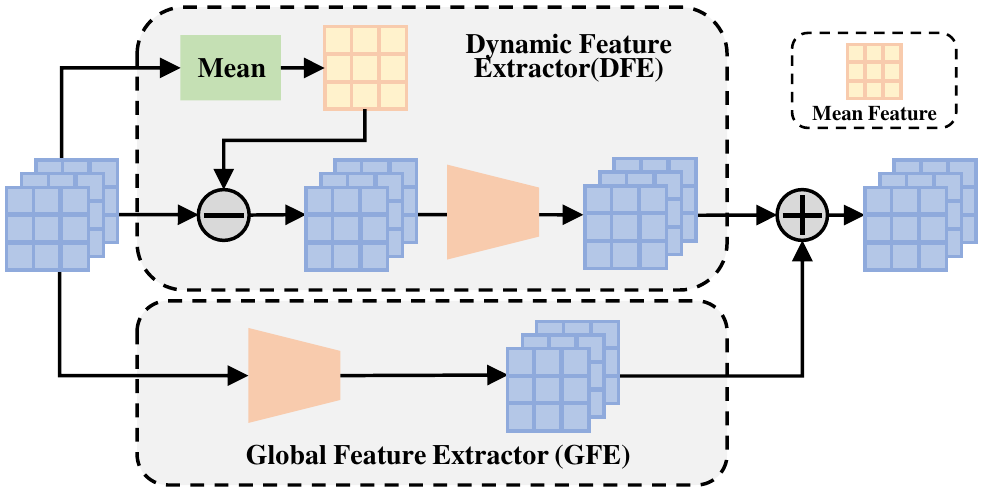}
\caption{The structure of Dynamic Augmentation Module (DAM).}
\vspace*{-1em}
\label{Fig_DAM}
\end{figure}

\subsection{Feature Extraction based on DAM}
Recently, many researchers use spatial-based \cite{wu2016comprehensive,shiraga2016geinet,zhang2016siamese,chao2019gaitset,zhang2019cross,zhang2019comprehensive,lin2020gait,chao2021gaitset,chai2021semantically,chai2021silhouette} or temporal-based models \cite{liao2017pose,PoseGait,lin2020gait,3D_pose,wolf2016multi,sokolova2019pose,thapar2019gait} to extract features for gait representations.
However, these methods do not pay enough attention to the dynamic information of the human body. As we mentioned above, the human torso and some disturbances, such as bags and coats, can be considered as static information, while moving limbs can be regarded as dynamic information. The bags and coats which do not belong to the human identification information may harm the recognition. On the other hand, the dynamic limbs often have larger changes than the relatively stable torso when walking, which indicates that the dynamic parts of the human body may provide more discriminative information. The traditional gait template based on mean function, such as Gait Energy Image (GEI) \cite{GEI}, preserves the information of the torso completely and weakens the role of dynamic limbs. To take advantage of dynamic information, we utilize the difference between the gait feature of each frame and the gait template based on the mean function to generate the dynamic feature map, the DAM block is shown in Figure \ref{Fig_DAM}. 

Assume that $X_{o} = \left \{ f_{i}|i = 1,2,...,T_{o} \right \} $, where $X_{o}  \in \mathbb{R}^{C_{o} \times T_{o} \times H_{o} \times W_{o}}$, $C_{o}$ is the number of input channels, $T_{o}$ is the length of the feature map and ($H_{o}$,$W_{o}$) is the image size of each frame. $f_{i}\in \mathbb{R}^{C_{o} \times 1 \times H_{o} \times W_{o}}$ is the $\textit{i}$-th frame of the feature map $X_{o}$. The dynamic feature map can be obtained by
\begin{equation}
X_{d} = \left \{ f_{i} - X_{m}|i = 1,2,...,T_{o} \right \},
\end{equation}
where
\begin{equation}
X_{m} = \frac{1}{T_{o}} \sum_{i=1}^{T_{o}}f_{i},
\end{equation}
$X_{d} \in \mathbb{R}^{C_{o} \times T_{o} \times H_{o} \times W_{o}}$ is the dynamic feature map, and $X_{m} \in \mathbb{R}^{C_{o} \times 1 \times H_{o} \times W_{o}}$ is the gait template based on mean function. Based on the dynamic feature map, we propose a Dynamic Feature Extractor (DFE) to establish the spatial-temporal feature representations of the dynamic parts of gait. The Dynamic Feature Extractor (DFE) can be designed as
\begin{equation}
Y_{DFE} =  C^{3 \times 3 \times 3} (X_{d}),
\end{equation}
where $Y_{DFE} \in \mathbb{R}^{C_{od} \times T_{o} \times H_{o} \times W_{o}}$ is the output of the DFE, $C_{od}$ is the number of output channels and $C^{3 \times 3 \times 3}$ denotes the 3D convolution operation with kernel size $(3, 3, 3)$.

Considering that DFE focuses on extracting the dynamic information of the human body, we add a Global Feature Extractor (GFE) to extract the global features of a gait sequence. The Global Feature Extractor (GFE) can be represented as
\begin{equation}
Y_{GFE} =  C^{1 \times 3 \times 3} (X_{o}),
\end{equation}
where $Y_{GFE} \in \mathbb{R}^{C_{od} \times T_{o} \times H_{o} \times W_{o}}$ is the output of the GFE and $C^{1 \times 3 \times 3}$ denotes the 3D convolution operation with kernel size $(1,3,3)$.

Based on the DFE, in this paper, we propose a novel module, DAM, to produce the dynamic augmentation features, greatly improving the representation ability. The DAM can be denoted as
\begin{equation}
Y_{DAM} =  \sigma ( Y_{GFE} + Y_{DFE} ),
\end{equation}
where $\sigma$ means LeakyRelu function. The Augmented Feature Maps (AFM) after DAM can be represented as
\begin{equation}
Y_{AFM} =  \sigma ( C^{3 \times 3 \times 3} (X_{o})) + Y_{DAM}
\end{equation}
where $Y_{AFM} \in \mathbb{R}^{C_{od} \times T_{o} \times H_{o} \times W_{o}}$.

As shown in Figure \ref{overview}, the feature extraction stage is implemented by multiple convolutions with DAM and the max-pooling operation.

\subsection{Loss Function}\label{loss_function}
To achieve the best performance, triplet loss and cross-entropy loss are used to train our network \cite{chao2021gaitset, fu2023sign}. Assume that $F_{i}$, $F_{j}$ and $F_{k}$ are feature representations corresponding to samples ${i}$, ${j}$ and ${k}$, respectively. Note that samples ${i}$ and ${j}$ are from class $A$, and the sample ${k}$ belongs to class $B$. The combined loss function can be represented as
\begin{equation}
\label{combined_loss}
Loss_{all} = Loss_{tri} + Loss_{cse},
\end{equation}
where $Loss_{tri}$ and $Loss_{cse}$ mean triplet loss and cross-entropy loss respectively.

On the one hand, the triplet loss $Loss_{tri}$ is proposed to optimize the inter-class and intra-class distances, which can be defined as
\begin{equation}
\label{tri}
Loss_{tri} = [D(F_{i}, F_{k}) - D(F_{i},F_{j}) + m]_{+},
\end{equation}
where $D(F_{i},F_{k})$ is the Euclidean distance between features of samples $i$ and $k$, $m$ means the margin of the triplet loss, and $[\alpha]_{+}$ is equal to $max(\alpha,0)$. 

On the other hand, the cross entropy loss $Loss_{cse}$ is introduced to optimize the classification space, which can be formulated as
\begin{equation}
Loss_{cse} = -\frac{1}{N}\sum_{i=1}^{N}log\frac{e^{W_{y_{i}}^{T}x_{i}+b_{y_{i}}}}{\sum_{j=1}^{n}e^{W_{j}^{T}x_{i}+b_{j}}}
\end{equation}
where $x_{i}$ is the feature of the $i$-th sample, and its label is $y_{i}$.

In our method, we obtain multiple column vectors at the Horizontal Mapping stage and then calculate the loss of each column vector following Equation \ref{combined_loss} \cite{lin2020gait,hou2020gait}. 

\subsection{Training Details and Test}
\label{Traintest}
\noindent \textbf{Training.} 
During the training phase, we first extract feature maps based on the dynamic augmentation model from the input sequence. Then, Temporal Aggregation (TA) and Horizontal Mapping (HM) are used to generate the fixed-size feature representation. After that, triplet loss and cross-entropy loss are used for evaluation. The sampling strategy is Batch ALL (BA) \cite{hermans2017defense,chao2019gaitset} and $P\times K$ instances are sampled in each step, where $P$ corresponds to the number of subject IDs, and $K$ denotes the number of samples for each subject ID.

\noindent \textbf{Test.} 
During the test phase, the whole input sequences can be fed into the proposed network to generate a feature representation $Y_{HM}$ that represents human gait. 
To evaluate the proposed method, we adopt the pattern ''Gallery-Probe'' to calculate Rank-1 accuracy. Therefore, the test set is split into two sets, \ie the gallery set and the probe set. Firstly, we input the gait sequence from all of the gallery set into the proposed network to generate feature representations, as the standard view sets. Secondly, each gait sequence of the probe set is put into the proposed network for feature representation. Then, this feature representation is used to calculate the Euclidean distance with all representations of the standard view sets. The label of the sample which has the smallest distance with the input sample is assigned to the input sample. Finally, we calculate the average accuracy to evaluate the performance of the proposed method.

\begin{table}[t]
  \centering
  \caption{Rank-1 accuracy (\%), Rank-5 accuracy (\%), Rank-10 accuracy (\%), and Rank-20 accuracy (\%) on the GREW dataset.}
  \vspace*{1em}
\renewcommand{\arraystretch}{1.2}
  \setlength{\tabcolsep}{1.5mm}{
    \begin{tabular}{c|c|c|c|c}
    \toprule
    Methods & \multicolumn{1}{c|}{Rank-1} & \multicolumn{1}{c|}{Rank-5} & \multicolumn{1}{c|}{Rank-10} & \multicolumn{1}{c}{Rank-20} \\
    \midrule
    PoseGait \cite{PoseGait} & 0.2   & 1.0  & 2.2  & 4.3  \\
    \hline
    GaitGraph \cite{GaitGraph} & 1.3   & 3.5  & 5.1  & 7.5  \\
    \hline
    GEINet \cite{shiraga2016geinet} & 6.8   & 13.4  & 17.0  & 21.0  \\
    \hline
    TS-CNN \cite{wu2016comprehensive} & 13.6  & 24.6  & 30.2  & 37.0  \\
    \hline
    GaitSet \cite{chao2019gaitset} & 46.3  & 63.6  & 70.3  & 76.8  \\
    \hline
    GaitPart \cite{fan2020gaitpart} & 44.0  & 60.7  & 67.3  & 73.5  \\
    \hline
    GaitGL \cite{lin2021gait} & 47.3  & 63.6  & 69.3  & 74.2  \\
    \hline
    MGN \cite{wang2018learning}  & 44.5  & 61.3  & 67.7  & 72.7  \\
    \hline
    
    CSTL \cite{huang2021context} & 50.6  & 65.9  & 71.9  & 76.9  \\
    \hline
    MTSGait \cite{zheng2022gaitmulti} & 55.3  & 71.3 & 76.9  & 81.6  \\
    \hline
    OpenGait \cite{fan2022opengait}  & 60.1  & -  & -  & -  \\
    \hline
    Ours  & \textbf{71.4} & \textbf{83.2} & \textbf{86.8} & \textbf{89.5} \\
    \bottomrule
    \end{tabular}%
    }
  \label{comparision_GREW}%
\end{table}%

\section{Experiments}

\subsection{Datasets}

To evaluate the performance of the proposed method, we conduct experiments on four popular gait datasets, including two real-word datasets \ie GREW \cite{zhu2021gait} and Gait3D \cite{zheng2022gait}, and two datasets captured from experimental environment, \ie CASIA-B \cite{yu2006framework} and OU-MVLP \cite{takemura2018multi}.

\noindent \textbf{GREW.} The GREW dataset \cite{zhu2021gait} is a large-scale outdoor gait dataset. The GREW includes 26,345 subjects and 128,671 sequences captured by 882 cameras in open environments, providing data in the form of Silhouettes, Optical Flow, GEI, and 2D/3D human poses. 
GREW has performed age grouping and gender annotation for all subjects. These subjects are divided into five age groups, with the adult groups (\ie, 16-30, 31-45, 46-60 years), the Children group (under 16 years) and the elder group (over 60 years). Each group has the males and females with similar amount. The GREW also provides five carrying conditions and six dressing styles.
The GREW is divided into three parts, \ie the training set with 20,000 identities and 102,887 sequences, the validation set with 345 identities and 1,784 sequences, and the test set with 6,000 identities and 24,000 sequences. These three sets of identities are captured by different cameras. Each test subject has four sequences, two of which are taken as probes and the other two as gallery. In addition, there is a distractor set with 233,857 sequences.

\noindent \textbf{Gait3D. }The Gait3D \cite{zheng2022gait} is a large-scale dataset, which contains 4,000 subjects and over 25,000 sequences captured from an unconstrained indoor scene by 39 cameras. It provides 3D SMPL models recovered from video. To facilitate comparison with other algorithms, the Gait3D dataset is divided into the training set with 3000 subjects and the test set with 1000 subjects. For the test set, the probe set of 1000 sequences is constructed by randomly selecting one sequence from each subject, and the rest sequences are put into the gallery set.

\noindent \textbf{CASIA-B.} The CASIA-B dataset \cite{yu2006framework} is one of the most popular gait databases, which consists of 124 subjects. For each subject, the CASIA-B dataset collected 10 groups of gait sequences (NM\#01-06, BG\#01-02 and CL\#01-02) under three conditions of normal walking (NM), walking with a bag (BG), and walking in coats (CL). Each group contains 11 videos from different view angles ($0^\circ$, $18^\circ$, ..., $162^\circ$, $180^\circ$). In experiments, methods usually adopt three protocols (Small-scale Training (ST), Medium-scale Training (MT), and Large-scale Training (LT)) to evaluate the performance of the proposed method \cite{chao2019gaitset}. For these three settings, 24, 62, and 74 subjects are taken as the training set and the rest 100, 62, and 50 subjects are used for test, respectively.
During the training stage, all gait sequences of the training set are used to train the network. In the test phase, gait sequences NM\#01-04 are taken as the gallery set, and gait sequences NM\#05-06, BG\#01-02 and CL\#01-02 are used as the probe set to calculate Rank-1 accuracy.


\noindent \textbf{OU-MVLP.} The OU-MVLP \cite{takemura2018multi} is one of the largest gait recognition datasets. It includes 10,307 subjects, each of which has two groups of gait sequences (Seq\#00 and Seq\#01). Each group collects 14 angles of gait sequences ($0^\circ$, $15^\circ$,..., $75^\circ$, $90^\circ$, $180^\circ$, $195^\circ$,..., $255^\circ$, $270^\circ$). In our experiments, we use gait sequences of 5,153 subjects to train the proposed network and take rest sequences as the test set to evaluate the performance \cite{chao2019gaitset}. During the test phase, sequences in Seq\#01 are defined as the gallery set, and sequences in Seq\#00 are considered as the probe set to calculate Rank-1 accuracy.


\begin{table}[htbp]
  \centering
  \caption{Rank-1 accuracy (\%), Rank-5 accuracy (\%), mAP (\%) and mINP on the Gait3D dataset.}
  \vspace*{1em}
\renewcommand{\arraystretch}{1.2}
  \setlength{\tabcolsep}{2.1mm}{
    \begin{tabular}{c|c|c|c|c}
    \toprule
    Methods & \multicolumn{1}{c|}{Rank-1} & \multicolumn{1}{c|}{Rank-5} & \multicolumn{1}{c|}{mAP} & \multicolumn{1}{c}{mINP}  \\
    \midrule
    PoseGait \cite{PoseGait} & 0.24  & 1.08  & 0.47  & 0.34  \\
    \hline
    GaitGraph \cite{teepe2021gaitgraph} & 6.25  & 16.23 & 5.18  & 2.42  \\
    \hline
    GEINet \cite{shiraga2016geinet} & 5.40  & 14.20  & 5.06  & 3.14   \\
    \hline
    GaitSet \cite{chao2019gaitset} & 36.70  & 58.30  & 30.01  & 17.30   \\
    \hline
    GaitPart \cite{fan2020gaitpart} & 28.20  & 47.60  & 21.58  & 12.36   \\
    \hline
    GLN \cite{hou2020gait}  & 31.40  & 52.90  & 24.74  & 13.58   \\
    \hline
    GaitGL \cite{lin2021gait}  & 29.70  & 48.50  & 22.29  & 13.26   \\
    \hline
    CSTL \cite{huang2021context}  & 11.70  & 19.20  & 5.59  & 2.59   \\
    \hline
    SMPLGait \cite{zheng2022gait} & 46.30  & 64.50  & 37.16  & 22.23   \\
    \hline
    MTSGait \cite{zheng2022gaitmulti} & 48.70  & 67.10  & 37.63  & 21.92  \\
    \hline
    OpenGait \cite{fan2022opengait} & 65.60  & -  & -  & -  \\
    \hline
    Ours  &  \textbf{66.30}    &   \textbf{80.80}    &   \textbf{56.40}    &  \textbf{37.30}  \\
    \bottomrule
    \end{tabular}%
    }
  \label{comparision_Gait3D}%
\end{table}%

{\subsection{Implementation Details}
We pre-process the original gait sequences. For GREW, CASIA-B and OU-MVLP, we normalize the size of each frame to $64 \times 44$. For Gait3D, the size of each frame is $128 \times 88$. 
In GREW and Gait3D, DyGait has five blocks built with the proposed DAM. For the CASIA-B and OU-MVLP, there are three DAM blocks to build the network.
In all experiments, $m$ in Equation \ref{tri} is set to 0.2.
In Section \ref{Traintest}, we introduce our sampling strategy Batch ALL (BA) in the training phase, which contains parameters $P$ and $K$. Parameters ($P$, $K$) are set to (32, 4) for GREW and Gait3D datasets. Parameters ($P$, $K$) are set to (8, 16) for the CASIA-B dataset. For the OU-MVLP dataset, $P$ and $K$ are set to 32 and 8, respectively.
SGD optimizer was adopted with the learning rate of 0.1 for GREW and Gait3D. For CASIA-B and OU-MVLP, Adam optimizer \cite{Adam} was adopted with the learning rate of 1e-4.
In the training stage, the frame length of each batch is set to 30.
The code of all experiments was written in PyTorch 1.1.0 \cite{PyTorch}. In the test phase, all frames of sequence can be taken as input to generate the feature representation. Iterations of the GREW, Gait3D, CASIA-B and OU-MVLP dataset are set to 200K, 150K, 80K and 210k, respectively.

\subsection{Comparison with State-of-the-art Methods}

\noindent \textbf{Evaluation on GREW.} We compare the performance of the proposed method with several gait recognition methods on the GREW dataset and show complete experimental results in Table \ref{comparision_GREW}. The comparison methods include PoseGait \cite{PoseGait}, GaitGraph  \cite{GaitGraph}, GEINeT \cite{GEINet}, TS-CNN  \cite{xing2016complete}, GaitSet \cite{chao2019gaitset}, GaitPart \cite{fan2020gaitpart}, GaitGL \cite{lin2021gait}, MGN \cite{wang2018learning}, CSTL \cite{huang2021context} , MTSGait \cite{zheng2022gaitmulti} and OpenGait \cite{fan2022opengait}. From Table \ref{comparision_GREW}, we find that gait recognition methods that perform well in laboratory scenarios degrade significantly on real scenario datasets. 
Although the real scenario dataset is subject to many external factors, our method performs 11.3\% higher than the state-of-art method OpenGait \cite{fan2022opengait} on Rank-1 accuracy.
Besides, our method gets 16.1\%, 11.9\%, 9.9\% and 7.9\% higher accuracy than the MTSGait on Rank-1, Rank-5, Rank-10 and Rank-20, respectively. The experimental results indicate that the proposed method obtains the highest average accuracy. This may be because the proposed method can learn more discriminative dynamic information. Besides, the methods \cite{PoseGait, GaitGraph} using the skeleton are less effective on real-world datasets, which may be that the skeleton-based approaches have less representation capability, the silhouette-based methods can obtain more discriminative feature representations.

\noindent \textbf{Evaluation on Gait3D.} Some competing gait recognition methods on the GREW dataset are used on Gait3D as well. Compared with the methods in Table \ref{comparision_Gait3D}, since our method pays more attention to the dynamic feature of the human body which obtains more specific information about legs and arms, our method achieves more appealing performance than the other gait recognition methods. Its accuracy is 0.7\% higher than the OpenGait \cite{fan2022opengait} on Rank-1. And the accuracy is 17.6\% and 13.7\% greater than MTSGait on Rank-1 and Rank-5, respectively. It shows that the captured dynamic features in our method may be more discriminative, which contributes to better performance.

\begin{table*}[t!]
  \scriptsize
  \centering
   \caption{Rank-1 accuracy (\%) on CASIA-B under all view angles and different conditions in LT setting, excluding identical-view case.
}
\vspace*{1em}
  \resizebox{0.99\textwidth}{!}{
\renewcommand{\arraystretch}{1.2}
    \begin{tabular}{c|c|c|c|c|c|c|c|c|c|c|c|c|c}
    \toprule
    \multicolumn{2}{c|}{Gallery NM\#1-4}  &\multicolumn{12}{c}{$0^{\circ}$-$180^{\circ}$} \\
    \hline
    \multicolumn{2}{c|}{Probe}    & $0^{\circ}$     & $18^{\circ}$    & $36^{\circ}$    & $54^{\circ}$    & $72^{\circ}$    & $90^{\circ}$    & $108^{\circ}$   & $126^{\circ}$   & $144^{\circ}$   & $162^{\circ}$   & $180^{\circ}$  & Mean\\
    \midrule

    \multicolumn{1}{c|}{\multirow{7}[2]{*}{NM\#5-6}} & GaitSet  & 90.8  & 97.9  & 99.4  & 96.9  & 93.6  & 91.7  & 95.0  & 97.8  & 98.9  & 96.8  & 85.8  & 95.0   \\
\cline{2-14}    & GaitPart  & 94.1  & 98.6  & 99.3  & 98.5  & 94.0  & 92.3  & 95.9  & 98.4  & 99.2  & 97.8  & 90.4  & 96.2   \\

\cline{2-14}    & MT3D  & 95.7  & 98.2  & 99.0  & 97.5  & 95.1  & 93.9  & 96.1  & 98.6  & 99.2  & 98.2  & 92.0  & 96.7   \\
\cline{2-14}    & GaitGL  & 96.0  & 98.3  & 99.0  & 97.9  & 96.9 & 95.4  & 97.0  & 98.9  & 99.3  & 98.8  & 94.0  & 97.4   \\
\cline{2-14}    & OpenGait  & -  & -  & -  & -  & -  & -  & -  & -  & -  & -  & -  & 97.6   \\
\cline{2-14}    & MetaGait  & 97.3  & \textbf{99.2}  & \textbf{99.5}  & \textbf{99.1}  & 97.2  & 95.5  & 97.6  & 99.1  & 99.3  & 99.1  & 96.7  & 98.1   \\
\cline{2-14}    & DyGait (ours)  & \textbf{97.4} & 98.9 & 99.2 & 98.3 & \textbf{97.7}  & \textbf{96.8} & \textbf{98.2} & \textbf{99.3} & \textbf{99.3} & \textbf{99.2} & \textbf{97.6} & \textbf{98.4}  \\
\hline
\hline
 \multicolumn{1}{c|}{\multirow{7}[2]{*}{BG\#1-2}} & GaitSet & 83.8  & 91.2  & 91.8  & 88.8  & 83.3  & 81.0  & 84.1  & 90.0  & 92.2  & 94.4  & 79.0  & 87.2   \\
\cline{2-14}  & GaitPart  & 89.1  & 94.8  & 96.7  & 95.1  & 88.3  & 84.9  & 89.0  & 93.5  & 96.1  & 93.8  & 85.8  & 91.5   \\
\cline{2-14}    & MT3D  & 91.0  & 95.4  & 97.5  & 94.2  & 92.3  & 86.9  & 91.2  & 95.6  & 97.3  & 96.4  & 86.6  & 93.0   \\
\cline{2-14}  & OpenGait  & -  & -  & -  & -  & -  & -  & -  & -  & -  & -  & -  & 94.0  \\
\cline{2-14}    & GaitGL  & 92.6  & 96.6  & 96.8  & 95.5  & 93.5 & 89.3  & 92.2  & 96.5  & 98.2 & 96.9  & 91.5  & 94.5   \\
\cline{2-14}    & MetaGait  & 92.9  & 96.7  & 97.1  & \textbf{96.4}  & 94.7 & 90.4  & 92.9  & 97.2  & 98.5 & \textbf{98.1}  & 92.3  & 95.2   \\
\cline{2-14}     & DyGait (ours)  & \textbf{94.5} & \textbf{96.9} & \textbf{97.4} & 96.1 & \textbf{95.4}  & \textbf{94.0} & 94.8 & \textbf{97.6} & \textbf{98.5}  & 97.7 & \textbf{94.9} & \textbf{96.2}  \\
\hline
\hline
 \multicolumn{1}{c|}{\multirow{7}[2]{*}{CL\#1-2}} & GaitSet & 61.4  & 75.4  & 80.7  & 77.3  & 72.1  & 70.1  & 71.5  & 73.5  & 73.5  & 68.4  & 50.0  & 70.4   \\
\cline{2-14}    & GaitPart  & 70.7  & 85.5  & 86.9  & 83.3  & 77.1  & 72.5  & 76.9  & 82.2  & 83.8  & 80.2  & 66.5  & 78.7   \\
\cline{2-14}    & OpenGait  & -  & -  & -  & -  & -  & -  & -  & -  & -  & -  & -  & 77.4   \\
\cline{2-14}   & MT3D & 76.0  & 87.6  & 89.8  & 85.0  & 81.2  & 75.7  & 81.0  & 84.5  & 85.4  & 82.2  & 68.1  & 81.5   \\
\cline{2-14}    & GaitGL  & 76.6  & 90.0  & 90.3  & 87.1  & 84.5  & 79.0  & 84.1  & 87.0  & 87.3  & 84.4  & 69.5  & 83.6   \\
\cline{2-14}    & MetaGait  & 80.0  & 91.8  & 93.0  & 87.8  & 86.5  & 82.9  & 85.2  & 90.0  & 90.8  & \textbf{89.3}  & \textbf{78.4}  & 86.9   \\
\cline{2-14}   & DyGait (ours)  & \textbf{82.2} & \textbf{93.0} & \textbf{95.2} & \textbf{91.6} & \textbf{87.1} & \textbf{83.4} & \textbf{87.2} & \textbf{90.1} & \textbf{92.4} & 88.2 & 75.8 & \textbf{87.8}
\\

    \bottomrule
    \end{tabular}%
 }
\label{comparision_casia}
\vspace*{0em}
\end{table*}%

\begin{table*}[htbp]
  \centering
  \caption{Rank-1 accuracy (\%) on OU-MVLP dataset under different view angles, excluding invalid probe sequences.}
  \vspace*{1em}
  \resizebox{0.99\textwidth}{!}{
\renewcommand{\arraystretch}{1.2}
    \begin{tabular}{cc|c|c|c|c|c|c|c|c|c|c|c|c|c|c}
    \toprule
    \multirow{2}[2]{*}{\textbf{Method}} & \multicolumn{14}{|c|}{\textbf{Probe View}}                                                            & \multicolumn{1}{c}{\multirow{2}[2]{*}{\textbf{Mean}}}  \\
\cline{2-15}    \multicolumn{1}{c|}{} & $0^{\circ}$ & $15^{\circ}$ & $30^{\circ}$ & $45^{\circ}$ & $60^{\circ}$ & $75^{\circ}$ & $90^{\circ}$ & $180^{\circ}$ & $195^{\circ}$ & $210^{\circ}$ & $225^{\circ}$ & $240^{\circ}$ & $255^{\circ}$ & $270^{\circ}$ &   \\
    \midrule
    
   

    \multicolumn{1}{c|}{GEINet } & 24.9  & 40.7  & 51.6  & 55.1  & 49.8  & 51.1  & 46.4  & 29.2  & 40.7  & 50.5  & 53.3  & 48.4  & 48.6  & 43.5  & 45.3   \\
    \hline
    \multicolumn{1}{c|}{GaitSet } & 84.5  & 93.3  & 96.7  & 96.6  & 93.5  & 95.3  & 94.2  & 87.0  & 92.5  & 96.0  & 96.0  & 93.0  & 94.3  & 92.7  & 93.3   \\
    \hline
    \multicolumn{1}{c|}{GaitPart } & 88.0  & 94.7  & 97.7  & 97.6  & 95.5  & 96.6  & 96.2  & 90.6  & 94.2  & 97.2  & 97.1  & 95.1  & 96.0  & 95.0  & 95.1   \\
    \hline
    \multicolumn{1}{c|}{GLN }   & 89.3  & 95.8  & 97.9  & 97.8  & 96.0  & 96.7  & 96.1  & 90.7  & 95.3  & 97.7  & 97.5  & 95.7  & 96.2  & 95.3  & 95.6   \\
    \hline
    \multicolumn{1}{c|}{SRN+CB} & 91.2  & 96.5  & 98.3  & 98.4  & 96.3  & 97.3  & 96.8  & 92.3  & 96.3  & 98.1  & 98.1  & 96.0  & 97.0  & 96.2  & 96.4   \\
    \hline
    \multicolumn{1}{c|}{GaitGL} & 90.5  & 96.1  & 98.0  & 98.1  & 97.0  & 97.6  & 97.1  & 94.2  & 94.9  & 97.4  & 97.4  & 95.7  & 96.5  & 95.7  & 96.2   \\
    \hline
 
    \multicolumn{1}{c|}{DyGait (ours)}  & \textbf{96.2} & \textbf{98.2} & \textbf{99.1} & \textbf{99.0} & \textbf{98.6} & \textbf{99.0} & \textbf{98.8} & \textbf{97.9} & \textbf{97.6} & \textbf{98.8} & \textbf{98.6} & \textbf{98.1} & \textbf{98.3} & \textbf{98.2} & \textbf{98.3} \\

    \bottomrule
    \end{tabular}%
}
  \label{comparision_oumvlp}%
      \vspace*{-1.5em}
\end{table*}%

\noindent \textbf{Evaluation on CASIA-B.} Our approach not only shows better performance on real scenario datasets, but also in lab scenarios. We compare the performance of the proposed method with several gait recognition methods on the CASIA-B dataset and show complete experimental results in Table \ref{comparision_casia}. Comparison methods include GaitSet \cite{chao2019gaitset}, GaitPart \cite{fan2020gaitpart}, MT3D \cite{lin2020gait},  GaitGL \cite{lin2021gait}, OpenGait \cite{fan2022opengait} and MetaGait \cite{dou2022metagait}.  Experimental results indicate that the proposed method has the highest average accuracy in all of the conditions (NM, BG and CL). We further explore the performance of the methods with different conditions.
It can be observed that the recognition accuracy has a significant decrease when the external environment changes. For example, recognition accuracies of GaitGL in NM, BG and CL are 97.4\%, 94.5\% and 83.6\% respectively. 
For the MetaGait framework, the corresponding accuracy is 98.1\%, 95.2\% and 86.9\%, respectively. Since both methods mentioned above equivalently extract the information of different regions of the human gait, which contains only static features, they are more vulnerable to suffer from the complex external environment.
For the proposed method, the accuracy in NM, BG and CL is 98.4\%, 96.2\% and 87.8\%, respectively, we pay more attention to the dynamic information so that more discriminative features may be captured. Specifically, the recognition accuracy of our method outperforms that of the other methods in BG and CL. Furthermore, our method performs well on some specific angles($0^{\circ}$ and $90^{\circ}$ ) in complex environment. For example, the accuracy of the MataGait is 92.9\% and 90.4\% in BG condition while that of our method is 94.5\% and 94.0\%.



\begin{figure}[t]
\centering
\includegraphics[width=0.46\textwidth]{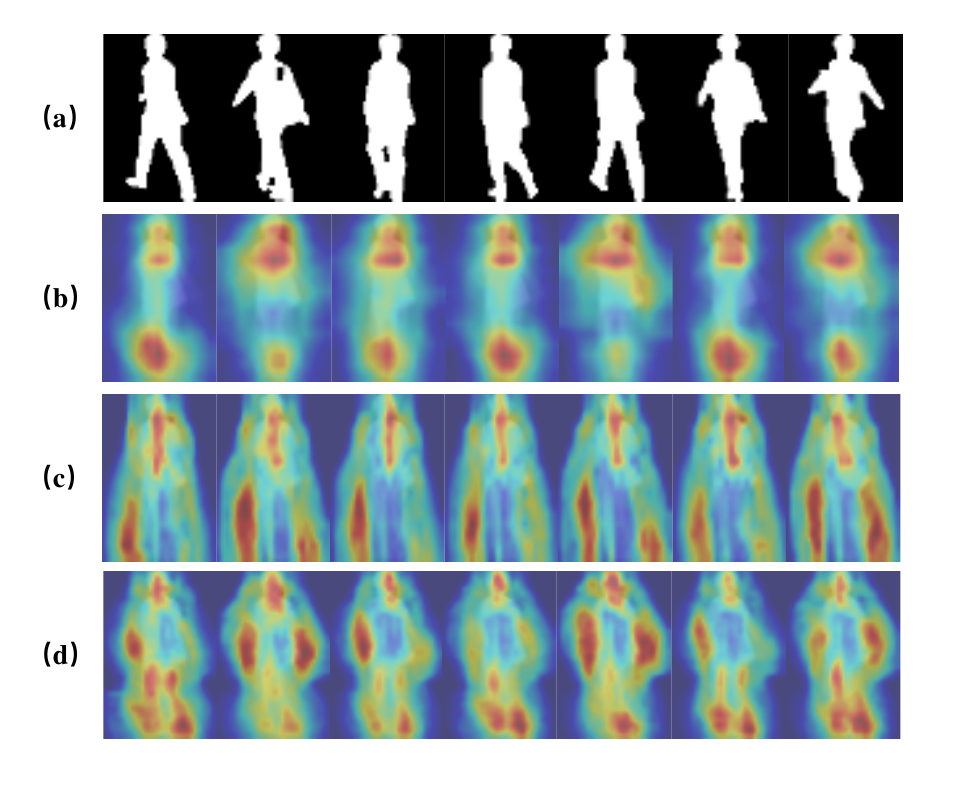}
\caption{(a) Original sequence. (b) Heatmaps for OpenGait~\cite{fan2022opengait}. (c) Heatmaps for GaitGL~\cite{lin2021gait}. (d) Heatmaps for our methods. 
}
\label{Fig_heatmap}
\end{figure}

\noindent \textbf{Evaluation on OU-MVLP.}
We compare the experimental result of our method with several state-of-the-art methods on the OU-MVLP dataset. The comparison methods include GEINet \cite{shiraga2016geinet}, GaitSet \cite{chao2019gaitset}, GaitPart \cite{fan2020gaitpart}, GLN \cite{hou2020gait},  SRN+CB \cite{hou2021set} and GaitGL \cite{lin2021gait}. Experimental results are shown in Table \ref{comparision_oumvlp}, which indicates that the proposed method achieves better recognition accuracy than the state-of-the-art methods. The accuracy of the proposed method is 98.3\% which outperforms the GaitGL by 2.1\%.
Furthermore, it can be observed that the accuracy of some specific view angles ($0^{\circ}$ and $180^{\circ}$) is far below the average accuracy. The main reason may be that the gait in these view angles have less information than the others. Table \ref{comparision_oumvlp} shows that the accuracy of our method at $0^{\circ}$ and $180^{\circ}$ is 96.2\% and 97.9\%, which outperforms GaitGL by 5.7\%, and 3.7\% respectively. This indicates that the dynamic feature extracted by our method can obtain more effective motion information, which is unique to each individual.

\begin{figure}[htbp]
\subfigure[Ours on CASIA-B]{
\centering
\includegraphics[width=0.23\textwidth]{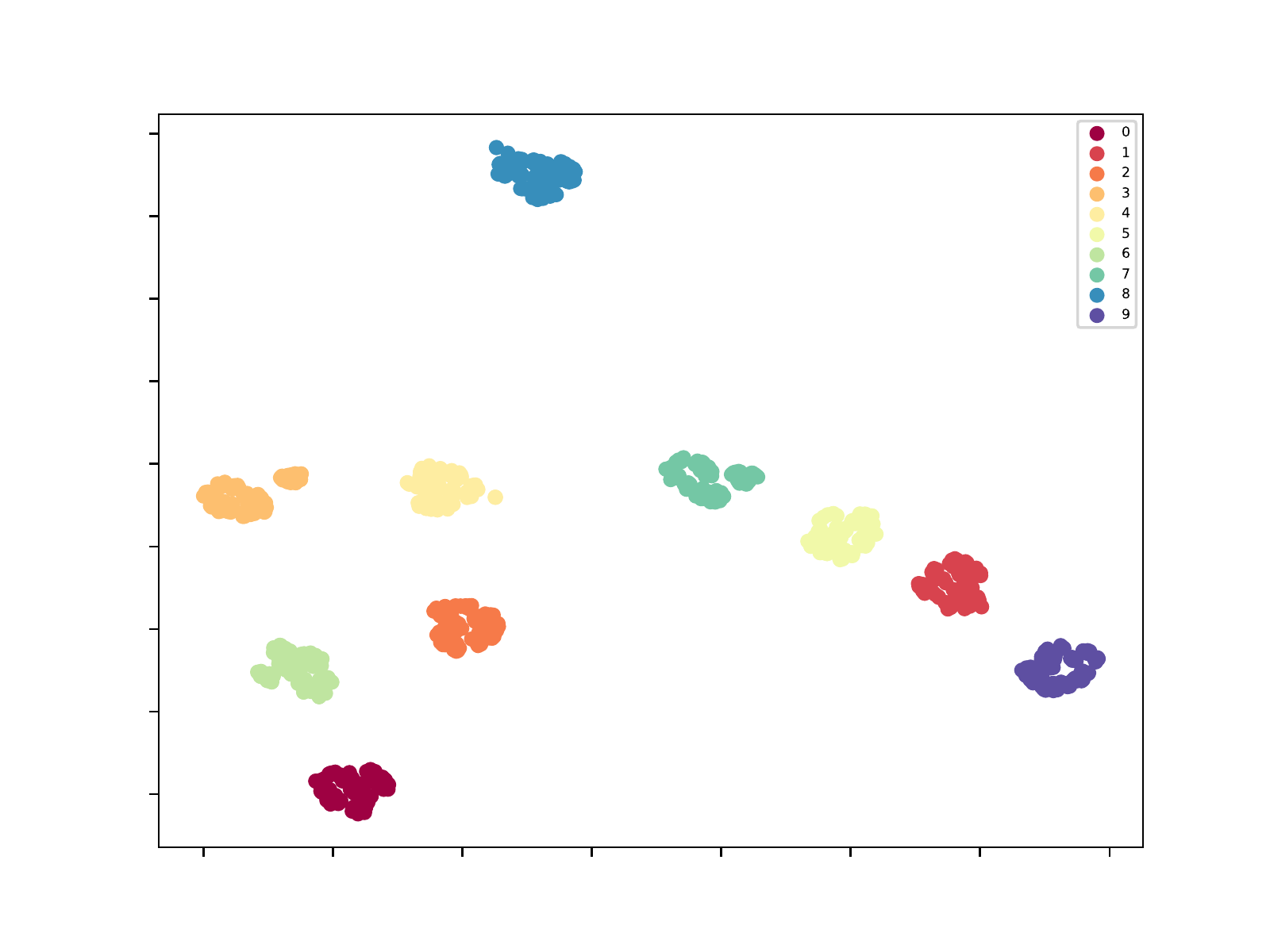}}
\subfigure[Ours w/o DFE on CASIA-B]{
\centering	
\includegraphics[width=0.23\textwidth]{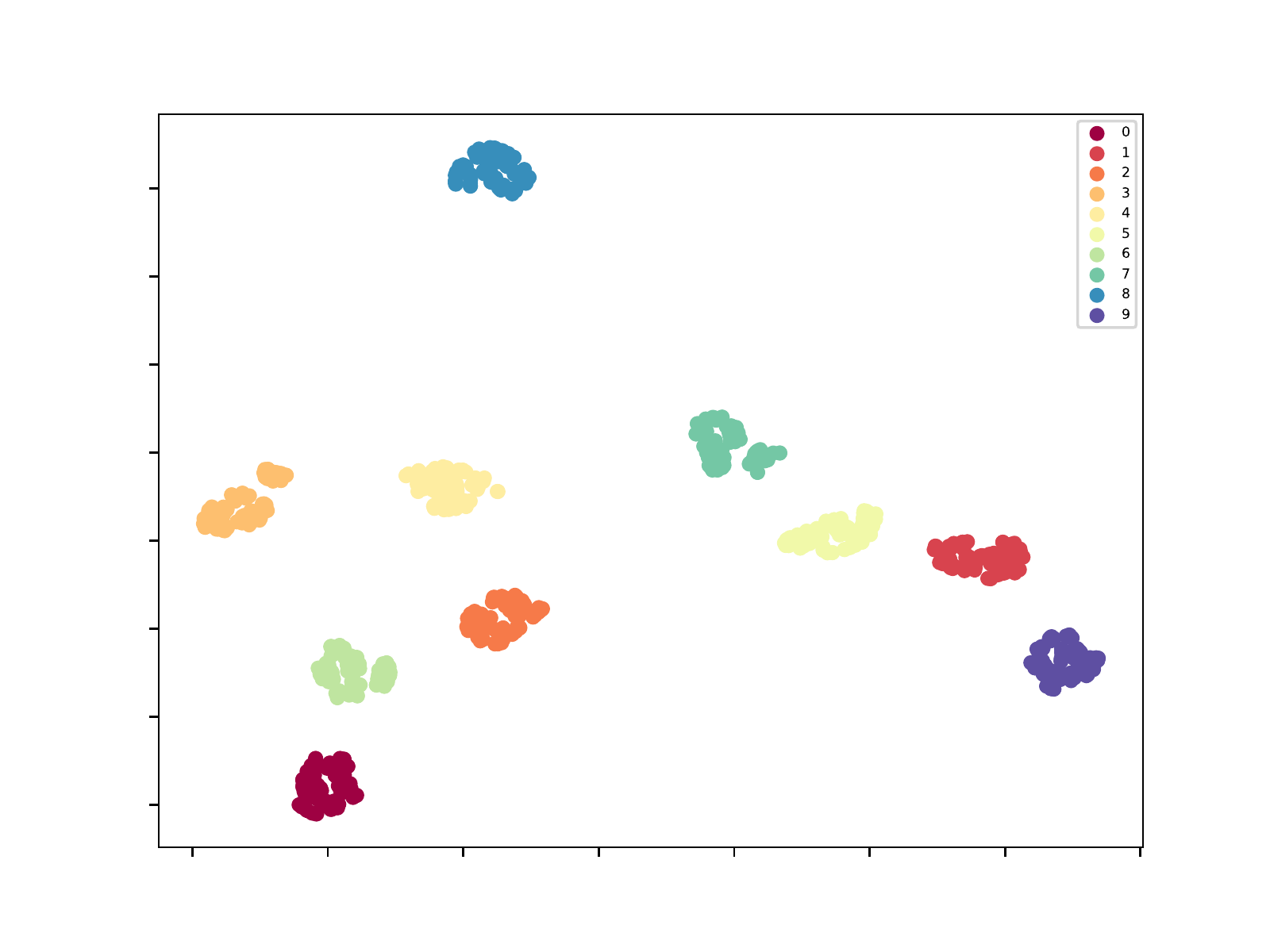}}
\caption{The visualization of feature distribution by t-SNE. (a) Our method on CASIA-B. (b) Our method without DAM.}
\vspace{-1em}
\label{Fig_vis}
\end{figure}

\subsection{Ablation Study} \label{Ablation_Study}
In Section \ref{method}, we propose the Dynamic Augmentation Module (DAM) to improve the feature representation ability. Therefore, we design more experiments to explore the role of these modules and some critical parameters.

\begin{table}[htbp]
  \centering
  \caption{The accuracy (\%) on GREW under different combinations of the proposed modules. GFE and DFE mean Global Feature Extractor and Dynamic Feature Extractor, respectively. }
  \vspace*{1em}
    \begin{tabular}{c|c|c|c|c|c}
    \toprule
    \multicolumn{2}{c|}{DAM} & \multirow{2}[2]{*}{Rank-1} & \multirow{2}[2]{*}{Rank-5} & \multirow{2}[2]{*}{Rank-10} & \multirow{2}[2]{*}{Rank-20} \\
\cline{1-2}    GFE   & DFE   &   &    &  \\
    \midrule
    \checkmark    & \checkmark     & \textbf{71.4}  & \textbf{83.2}  & \textbf{86.8}  & \textbf{89.5} \\
    \hline
    \checkmark    & $\times$       & 67.4  & 80.3  & 84.6 & 87.5    \\
    \hline
    $\times$      & \checkmark     & 70.0  & 82.2  & 85.9 & 88.7   \\
    
    \bottomrule
    \end{tabular}%
    \vspace*{0em}
  \label{table_Ablation_DAM}%
\end{table}%

\noindent \textbf{Analysis of DAM.}
To explore the contribution of the two branches used for feature extraction, we design another two models with only a single branch and conduct the experiments on the real scenario dataset GREW. The results are shown in Table \ref{table_Ablation_DAM}. It can be observed that the models with DFE achieve better performance than those without DFE. This is because DFE can enable the model to focus on the dynamic information of gait, which is helpful for recognition. It can be also observed that combining global features and dynamic features can bring extra performance gain. From the results, we can see that the accuracy of the model combining both the DFE and GFE performs 4.0\% better than the model without DFE. This indicates using the both branches to extract features can lead to more powerful and discriminative representation ability.



\noindent \textbf{Analysis of the number of DAMs.} 
In this paper, we propose a novel dynamic feature extractor to generate discriminative feature representations. The proposed dynamic feature extractor can be used to replace any layer of the network. To explore the optimal number of DAMs, we design the ablation studies by using different numbers of DAMs. The ablation studies are built on GREW. Experimental results are shown in Table \ref{tab_numberofdam}. 
It can be observed that higher recognition accuracy can be obtained by using a larger number of DAMs. 
Thereby, the number of DAMs on the GREW dataset is finally set to five.

\begin{table}[htbp]
  \centering
  \vspace*{0em}
  \caption{Rank-1 accuracy (\%) of different DAM number.}
  \vspace*{1em}
    \begin{tabular}{c|c|c|c|c|c}
    \toprule
    \multicolumn{2}{c|}{\multirow{2}[2]{*}{DAM}} & \multirow{2}[2]{*}{Rank-1} & \multirow{2}[2]{*}{Rank-5} & \multirow{2}[2]{*}{Rank-10} & \multirow{2}[2]{*}{Rank-20} \\
    \multicolumn{2}{c|}{} &       &       &  \\
    \midrule
    
    \multicolumn{2}{c|}{1} & 13.9  & 24.5  & 30.1 & 35.7 \\
    \hline
    \multicolumn{2}{c|}{2} & 41.4 & 56.8  & 62.8 & 68.0  \\
    \hline
    \multicolumn{2}{c|}{3} & 57.2  & 70.9 & 75.6 & 79.6  \\
    \hline
    \multicolumn{2}{c|}{4} & 69.2  & 80.8 & 84.7 & 87.9 \\
    \hline
    \multicolumn{2}{c|}{5} & \textbf{71.4} & \textbf{83.2}  & \textbf{86.8} & \textbf{89.5}\\
    \bottomrule
    \end{tabular}%
    \vspace*{-1em}
  \label{tab_numberofdam}%
\end{table}%

\section{Visualization}
In this section, we visualize the feature distribution of the models with DFE and withou DFE on CASIA-B dataset.
As shown in Figure \ref{Fig_vis}, we can observe that the intra-class distance of the features is closer and the inter-class distance is farther the DFE modules is added. The visualization demonstrates that by introducing the DFE which extracts the dynamic information, more discriminative representation can be obtained and contribute to better inter-class and intra-class distribution.

Furthermore, we visualize the heatmaps of the existing methods and our DyGait. As shown in Figure \ref{Fig_heatmap}, the methods \cite{chao2019gaitset,fan2022opengait} fail to distinguish the differences between the dynamic parts (\eg legs and arms) and the static parts (\eg torso) during walking. Figure \ref{Fig_heatmap} (d) shows our DyGait pays more attention to dynamic parts. This is because our DyGait explicitly models the motion information by DAM, which enhances the discriminativeness of moving parts.

\section{Conclusion}
In this paper, we propose a novel network to generate more discriminative  representations for gait recognition. The proposed DyGait includes both the Global Feature Extractor (GFE) and the Dynamic Feature Extractor (DFE) modules. The model can extract not only spatial-temporal features but also dynamic features from gait sequences. We develop the feature extraction module based on dynamic augmentation to generate augmented features. Taking the dynamic feature extractor as an extra branch can effectively enhance the discriminability of the global feature representation. The experiments on four popular datasets indicate that the proposed method can achieve optimal performance.

{\small
\bibliographystyle{ieee_fullname}
\bibliography{egbib}
}

\end{document}